\documentclass[11pt, oneside]{article}   	
\usepackage{geometry}                		
\usepackage{graphicx}				
\usepackage{amssymb}
\usepackage{amsmath}

\title{A Modified Construction for a Support Vector Machine to Accommodate Class Imbalances}
\author{Matt Parker, Colin Parker}
\date{}
\begin{document}
\maketitle

\begin{abstract}
   Given a training set with binary classification, the Support Vector Machine identifies the hyperplane maximizing the margin between the two classes of training data. This general formulation is useful in that it can be applied without regard to variance differences between the classes. Ignoring these differences is not optimal, however, as the general SVM will give the class with lower variance an unjustifiably wide berth. This increases the chance of misclassification of the other class and results in an overall loss of predictive performance. An alternate construction is proposed in which the margins of the separating hyperplane are different for each class, each proportional to the standard deviation of its class along the direction perpendicular to the hyperplane. The construction agrees with the SVM in the case of equal class variances. This paper will then examine the impact to the dual representation of the modified constraint equations.
\end{abstract}

\section{A Recap: The Classical SVM Construction}

For Section 1, we follow the construction given by Hastie, Tibshirani, and Freidman in \textit{The Elements of Statistical Learning} \cite{ESL}. We will parallel this approach in Section 2 when constructing the alternate method.\\

Suppose we have training data consisting of pairs of observations and labels, $(x_{i}, y_{i})$, for $i = 1, ... , N,$ with $x_{i} \in \mathbb{R}^{p}$ and $y_{i} \in \{-1, 1\}$. We may define a hyperplane by:

\begin{equation} \{x : f(x) = x^{T}\beta + \beta_{0} = 0\} \end{equation}

where $\beta$ is a vector perpendicular to the hyperplane. An associated classification rule is induced by:

\begin{equation} G(x) = \textrm{sign}[x^{T}\beta + \beta_{0}] \end{equation}

The goal of finding a separating hyperplane which maximizes the margin $M$ for a linearly separable dataset, the minimum perpendicular distance to a datapoint of either class, can be formalized as:

\begin{align}
& \max_{\beta, \beta_{0}, \|\beta\| = 1} M \\
& \textrm{subject to } y_{i}(x_{i}^{T}\beta + \beta_{0}) \geq M \,\,\,\, i = 1, ... , N
\end{align}

This can be more conveniently rephrased by removing the requirement $\beta$ be a unit vector, and setting $M = \frac{1}{\|\beta\|}$:

\begin{align}
& \min_{\beta, \beta_{0}} \|\beta\| \\
& \textrm{subject to } y_{i}(x_{i}^{T}\beta + \beta_{0}) \geq 1 \,\,\,\, i = 1, ... , N
\end{align}

Now define slack variables $\zeta_{i}, i = 1, ... , N$ by

\begin{equation} \zeta_{i}\,\, = \,\, \max \,\,(0, \,1 - y_{i}(x_{i}^{T}\beta + \beta_{0})) \end{equation}

This gives us a framework to relax the assumption of linear separability. Noting that misclassifications occur when $\zeta_{i} > 1$, we see the slack variables are the proportion of the margin by which various points fall within their respective margins. We may control the amount of slack by imposing the additional condition:

\begin{equation} \sum_{i=1}^{N}\zeta_{i} \leq \textrm{constant} \end{equation}

for some constant. This is computationally equivalent to the following expression:

\begin{align}
& \min_{\beta, \beta_{0}}\frac{1}{2}\|\beta\|^{2} + C\sum_{i=1}^{N}\zeta_{i} \\
& \textrm{subject to } \,\, \zeta_{i} \geq 0, \,\,\, y_{i}(x_{i}^{T}\beta + \beta_{0}) \geq 1-\zeta_{i} \,\,\, \forall i
\end{align}

where the parameter $C$ replaces the constant in the previous expression. The corresponding Lagrange primal function is given by:

\begin{equation}
L_{P} = \frac{1}{2}\|\beta\|^{2} + C\sum_{i=1}^{N}\zeta_{i} - \sum_{i=1}^{N}\alpha_{i}[y_{i}(x_{i}^{T}\beta + \beta_{0}) - (1 - \zeta_{i})] - \sum_{i=1}^{N}\mu_{i}\zeta_{i}
\end{equation}

which is to be minimized with respect to $\beta, \beta_{0}$, and $\zeta_{i}$. Setting the respective derivatives equal to zero, we get the equations:

\begin{align}
\beta & = \sum_{i=1}^{N}\alpha_{i}y_{i}x_{i} \\
0  & = \sum_{i=1}^{N} \alpha_{i}y_{i} \\
\alpha_{i} & = C - \mu_{i} \,\, \forall i
\end{align}

and positivity constraints $\alpha_{i}, \mu_{i}, \zeta_{i} \geq 0 \forall i$. By substituting the above three equations into the Lagrangian dual we obtain the Wolfe dual, given by:

\begin{align}
L_{D} & = \sum_{i=1}^{N}\alpha_{i} - \frac{1}{2}\sum_{i=1}^{N}\sum_{i'=1}^{N}\alpha_{i}\alpha_{i'}y_{i}y_{i'}x_{i}^{T}x_{i'} \\
& = \sum_{i=1}^{N}\alpha_{i} - \frac{1}{2}\sum_{i=1}^{N}\sum_{i'=1}^{N}\alpha_{i}\alpha_{i'}y_{i}y_{i'}\langle x_{i}, x_{i'}\rangle
\end{align}

In addition, the Karush-Kuhn-Tucker conditions yield:

\begin{align}
\alpha_{i}[y_{i}(x_{i}^{T}\beta + \beta_{0} - (1 - \zeta_{i})] & = 0 \\
\mu_{i}\zeta_{i} & = 0 \\
y_{i}(x_{i}^{T}\beta + \beta_{0}) - (1 - \zeta_{i}) & \geq 0
\end{align}

for $i = 1, ... , N$. These equations collectively uniquely define the solution to the dual problem.

\section{A Modified Approach: Accommodating Difference in Class Variance}

The original construction of the SVM for linearly separable data has the goal of maximizing the margin $M = \frac{1}{\|\beta\|}$. In the event of a noticeable difference between class variances in the direction of $\beta$ (perpendicular to our separating hyperplane), the SVM ends up positioning the decision boundary closer to the class with larger variance [say, class A] than would be optimal. The new construction accommodates these class imbalances by increasing the margin of the class of greater variance. \\

It will be useful at this point to define a few terms. For class $K$, element $x_{j} \in K$, and separating hyperplane $\{x : x^{T}\beta + \beta_{0} = 0\}$, define $\sigma_{K, \beta} = \sigma_{y_{j}, \beta}$ to be the standard deviation of elements of class $K$ in the direction of $\beta$:

\begin{align} \sigma_{K, \beta} \,\,=\,\, \sigma_{y_{j}, \beta} & = Var(\{(x_{i}-\overline{x})\cdot\left(\frac{\beta}{\|\beta\|}\right) \,\, \vert \,\, i \in K\})^{\frac{1}{2}}  \\
& = \left(\sum_{j : y_{j} = y_{i}}\left[\left(x_{j} - \overline{x}\right)\cdot\left(\frac{\beta}{\|\beta\|}\right)\right]^{2}\right)^{\frac{1}{2}}\end{align}

and, for class $K$ and arbitrary hyperplane $\{x : x^{T}\beta + \beta_{0} = 0\}$, define the margin of class $K$ to be:

\begin{equation} M_{K} = \min_{x_{i} \in K} \,\, y_{i}\left(\frac{x_{i}^{T}\beta + \beta_{0}}{\sigma_{y_{i}, \beta}}\right)\end{equation}

We will now seek to find the separating hyperplane which maximizes $\min_{K} M_{K}$, the minimum margin over all classes. As an aside, a byproduct of the classic construction of the SVM yields the equality $M_{A} = M_{B}$ when separating classes $A$ and $B$, since the maximum margin is obtained when the separating hyperplane is midway between both classes. Our new construction will yield as a byproduct the equality:

\begin{equation} \frac{M_{A}}{\sigma_{A, \beta}} \,\, = \,\, \frac{M_{B}}{\sigma_{B, \beta}} \end{equation}

This shows that in the event our classes have equal variance in the direction of $\beta$, the modified construction coincides with the classical SVM. \\

\section{Examining Implications to Dual Representation}

Maximizing $\min_{K}M_{K}$ modifies the optimization problem to the pair of equations:

\begin{align}
& \min_{\beta, \beta_{0}} \|\beta\| \\
& \textrm{subject to } y_{i}\left(\frac{x_{i}^{T}\beta + \beta_{0}}{\sigma_{y_{i}, \beta}}\right) \geq 1 \,\,\,\, i = 1, ... , N
\end{align}

Slightly redefining slack variables according to the fraction of the respective margins they span yields:

\begin{equation} \zeta_{i}\,\, = \,\, \max\left(0, \,\,1 - y_{i}\left(\frac{x_{i}^{T}\beta + \beta_{0}}{\sigma_{y_{i}, \beta}}\right)\right) \end{equation}

and the corresponding modified SVM equations are given by:

\begin{align}
& \min_{\beta, \beta_{0}}\frac{1}{2}\|\beta\|^{2} + C\sum_{i=1}^{N}\zeta_{i} \\
& \textrm{subject to } \,\, \zeta_{i} \geq 0, \,\,\, y_{i}\left(\frac{x_{i}^{T}\beta + \beta_{0}}{\sigma_{y_{i}, \beta}}\right) \geq 1-\zeta_{i} \,\,\,\,\,\,\,\,\,\forall i
\end{align}

We can now formulate the corresponding Lagrangian (primal) function as:

\begin{equation} L_{P} = \frac{1}{2}\|\beta\|^{2} + C\sum_{i=1}^{N}\zeta_{i} - \sum_{i=1}^{N}\alpha_{i}\left[y_{i}\sigma_{y_{i}, \beta}^{-1}(x_{i}^{T}\beta + \beta_{0}) - (1-\zeta_{i})\right] - \sum_{i=1}^{N}\mu_{i}\zeta_{i} \end{equation}

which we again minimize with respect to $\beta, \beta_{0}$, and $\zeta_{i}$. Setting derivatives with respect to $\beta_{0}$ and $\zeta_{i}$ equal to zero, we get similar results: 

\begin{align}
0  & = \sum_{i=1}^{N} \alpha_{i}y_{i}\sigma_{y_{i}, \beta}^{-1} \\
\alpha_{i} & = C - \mu_{i} \,\, \forall i
\end{align}

and a slightly more complex equation when doing the same with respect to $\beta$:

\begin{align}
0 & = \nabla_{\beta} L_{P} \\
& = \nabla_{\beta} \left(\frac{1}{2}\|\beta\|^{2} - \sum_{i=1}^{N}\left(\alpha_{i}y_{i}\right)\left(\sigma_{y_{i}, \beta}^{-1}\right)\left(x_{i}^{T}\beta + \beta_{0}\right)\right) \\
& = \beta - \sum_{i=1}^{N}\alpha_{i}y_{i}x_{i}\sigma_{y_{i}, \beta}^{-1} + \sum_{i=1}^{N}(\alpha_{i}y_{i})\left(\sigma_{y_{i}, \beta}^{-2}\right)\left(x_{i}^{T}\beta + \beta_{0}\right)\left(\nabla_{\beta}\sigma_{y_{i}, \beta}\right)
\end{align}

Expanding $\sigma_{y_{i}, \beta}$ to its representation in (21), we may utilize the Hadamard product notation $\circ$ and the fact 

\begin{equation}\nabla_{\beta}\left(\left(x_{j} - \overline{x}\right)\cdot\left(\frac{\beta}{\|\beta\|}\right)\right) = \left(x_{j} - \overline{x}\right)\cdot\left(\frac{\|\beta\|^{2} - \beta\circ\beta}{\|\beta\|^{3}}\right) \end{equation}

where $\circ$ is the Hadamard product, to obtain:

\begin{multline}
0 = \beta - \sum_{i=1}^{N}\alpha_{i}y_{i}x_{i}\sigma_{y_{i}, \beta}^{-1} \,\,\,+ \\
+ \sum_{i=1}^{N}\left[\alpha_{i}y_{i}\sigma_{y_{i}, \beta}^{-3}\left(x_{i}^{T}\beta + \beta_{0}\right)\left(\sum_{j : y_{j} = y_{i}}\left[\left(x_{j}-\overline{x}\right)\cdot\left(\frac{\beta}{\|\beta\|}\right)\right]\left[\left(x_{j}-\overline{x}\right)\left(\frac{\overrightarrow{\mathbf{1}}\|\beta\|^{2}-\beta\circ\beta}{\|\beta\|^{3}}\right)\right]\right)\right]
\end{multline}

where $\overrightarrow{\mathbf{1}}$ is the vector of ones [1, ... , 1]. \\

This gives us a working representation of the equivalent dual optimization equations under the new construction, and a forthcoming paper will be examining the solvability of the above in general in light of the other constraint equations, as well as consequent impacts to kernelizability of the method. We will also examine in depth the circumstances in which our alternate construction outperforms a traditional Support Vector Classifier, and attempt to quantify them. \\
\\


\begin{thebibliography}{9}
\bibitem{ESL} 
Trevor Hastie, Robert Tibshirani, and Jerome Freidman. 
\textit{The Elements of Statistical Learning}. 
Springer-Verlag, New York, New York, 2009.
 
\bibitem{Andrew_Ng} 
Andrew Ng. 
\textit{CS229 Lecture Notes}. 
[\texttt{http://cs229.stanford.edu/notes/cs229-notes3.pdf}] 

 
\bibitem{ESL} 
Robert Gunn,
\textit{Support Vector Machines for Classification and Regression}. 
Technical Report for University of Southampton, Southampton, England, 1998.
\end{thebibliography}
\end{document}